%% file: main.tex
\documentclass[conference]{IEEEtran}
\IEEEoverridecommandlockouts
\usepackage{amsmath,amssymb,amsfonts}
\usepackage{algorithmic}
\usepackage{graphicx}
\usepackage{textcomp}
\usepackage{xcolor}
\def\BibTeX{{\rm B\kern-.05em{\sc i\kern-.025em b}\kern-.08em
    T\kern-.1667em\lower.7ex\hbox{E}\kern-.125emX}}
    
\usepackage[numbers,sort&compress]{natbib}  
\usepackage{url}
\usepackage{booktabs}
\usepackage{mathtools}
\usepackage{stmaryrd}
\usepackage{circledsteps}
\usepackage{listings}

\usepackage[caption=false]{subfig}

\newcommand{\todo}[1]{{\leavevmode\color{red}#1}}
\newcommand{\remove}[1]{}

\begin{document}
\title{
Correlation-based Discovery of Disease Patterns for Syndromic Surveillance
}

\remove{
\author{\IEEEauthorblockN{
Anonymous}
\IEEEauthorblockA{
\textit{\ldots}\\
\ldots \\
\ldots}
\and
\IEEEauthorblockN{
Anonymous}
\IEEEauthorblockA{
\textit{\ldots}\\
\ldots \\
\ldots}
\and
\IEEEauthorblockN{
Anonymous}
\IEEEauthorblockA{
\textit{\ldots}\\
\ldots \\
\ldots}
\and
\IEEEauthorblockN{
Anonymous}
\IEEEauthorblockA{
\textit{\ldots}\\
\ldots \\
\ldots}
}
}

\author{\IEEEauthorblockN{
Michael Rapp}
\IEEEauthorblockA{
\textit{TU Darmstadt}\\
Darmstadt, Germany \\
0000-0001-8570-8240}
\and
\IEEEauthorblockN{
Moritz Kulessa}
\IEEEauthorblockA{
\textit{TU Darmstadt}\\
Darmstadt, Germany \\
0000-0001-8191-4485}
\and
\IEEEauthorblockN{
Eneldo Loza Menc\'{i}a}
\IEEEauthorblockA{
\textit{TU Darmstadt}\\
Darmstadt, Germany \\
0000-0002-2735-9326}
\and
\IEEEauthorblockN{
Johannes F\"urnkranz}
\IEEEauthorblockA{
\textit{JKU Linz}\\
Linz, Austria \\
0000-0002-1207-0159}
}

\maketitle

\begin{abstract}

Early outbreak detection is a key aspect in the containment of infectious diseases, as it enables the identification and isolation of infected individuals before the disease can spread to a larger population. Instead of detecting unexpected increases of infections by monitoring confirmed cases, syndromic surveillance aims at the detection of cases with early symptoms, which allows a more timely disclosure of outbreaks. However, the definition of these disease patterns is often challenging, as early symptoms are usually shared among many diseases and a particular disease can have several clinical pictures in the early phase of an infection. To support epidemiologists in the process of defining reliable disease patterns, we present a novel, data-driven approach to discover such patterns in historic data. The key idea is to take into account the correlation between indicators in a health-related data source and the reported number of infections in the respective geographic region. In an experimental evaluation, we use data from several emergency departments to discover disease patterns for three infectious diseases. Our results suggest that the proposed approach is able to find patterns that correlate with the reported infections and often identifies indicators that are related to the respective diseases.


\end{abstract}


\input{sections/1_introduction}

\input{sections/2_preliminaries}

\input{sections/3_syndrome_learning}
\input{sections/4_evaluation}

\input{sections/5_discussion}

\input{sections/6_conclusion}

\input{sections/6_acknowledgments}

\bibliographystyle{abbrvnat}
\footnotesize
\bibliography{main}


\end{document}

%% file: sections/1_introduction.tex
\section{Introduction}


Throughout history, major outbreaks of infectious diseases have caused millions of deaths and therefore pose a serious threat to public health. Among the most well-known outbreaks is the \emph{Great Influenza Pandemic} between the years $1918$ and $1920$, which has killed approximately $40$ million people worldwide, as well as the recent, still ongoing pandemic of \emph{SARS-CoV-2}~\citep{barro2020}. A fundamental strategy to diminish or even prevent the spreading of infectious diseases is to detect local outbreaks as early as possible in order to identify and isolate infected individuals.
For the early detection of unexpected increases in the number of infections, which may be an indicator for an outbreak, infectious diseases are under constant surveillance by epidemiologists.


Besides tracking the number of confirmed infections based on laboratory testing, a promising approach to outbreak detection is \emph{syndromic surveillance}~\citep{henning2004}, which focuses on monitoring the number of cases with early symptoms. Compared to laboratory testing, which can take several days until results are available, it allows for a more timely detection of outbreaks. Moreover, a much larger population can be put under surveillance by using health-related data sources that do not depend on confirmed results.
For example, the number of antipyretic drug sales in pharmacies could be considered as an indicator for an outbreak of influenza. Or, based on data that is gathered in emergency departments, the number of patients with a fever could serve as another indicator for this particular disease.

One of the major challenges in syndromic surveillance is the definition of such indicators, also referred to as \emph{syndromes} or \emph{disease patterns}. They highly depend on the infectious disease and the data source under surveillance. 
Since early symptoms are usually shared among many diseases and because a particular disease can have several clinical pictures at early stages of an infection, it is difficult to obtain reliable syndromes.
For this reason, the definition of disease patterns is usually based solely on expert knowledge of epidemiologists, a time-consuming and laborious process~\citep{mandl2004}.
This motivates the demand for tools that allow for a user-guided generation and comparison of syndrome definitions. To be useful in practice, such tools should be flexible enough to be applied to different types of data~\citep{hopkins2017}.


In this work, we present a data-driven approach that aims at supporting epidemiologists in the process of identifying 
disease patterns for infectious diseases.
It discovers syndrome definitions 
from health-related data sources, based on their correlation to the reported number of infections in the respective geographical area. As the first contribution of this work, we introduce a formal definition of this correlation-based discovery task. Our second contribution is an algorithm for the automatic extraction of disease patterns that utilizes techniques from the field of inductive rule learning. To provide insight into the data, the syndromes it discovers may be suggested to epidemiologists, who can adjust the input or the parameters of the algorithm to interactively refine the syndromes according to their domain knowledge. To better understand the capabilities and shortcomings of the proposed method, we evaluate its ability to reconstruct randomly generated disease patterns with varying characteristics. Furthermore, we apply our approach to emergency department data to learn disease patterns for Influenza, Norovirus and SARS-CoV-2. To assess the quality of the obtained patterns, we discuss the indicators they are based on and relate them to the number of infections according to publicly available reports, as well as handcrafted syndrome definitions.

\remove{

In this work, we present a data-driven approach which can support epidemiologists during that process by
In this work, we present a data-driven approach based on rule learning which can support epidemiologists to define syndromes. \ldots

In this work, we present a data-driven approach which can support epidemiologists to define syndromes. Viewed from a machine learning perspective

a rule-based approach to support epidemiologists

\begin{itemize}
    \item Importance outbreak detection
    \item Traditional surveillance
    \item Syndromic surveillance
    \item Major challenge to find suitable syndromes
    \item Learning of Syndromes
    \item Refinement of syndrome definitions
\end{itemize}
}

%% file: sections/2_preliminaries.tex
\begin{figure*}[t]
    \centering
    \includegraphics[width=\textwidth]{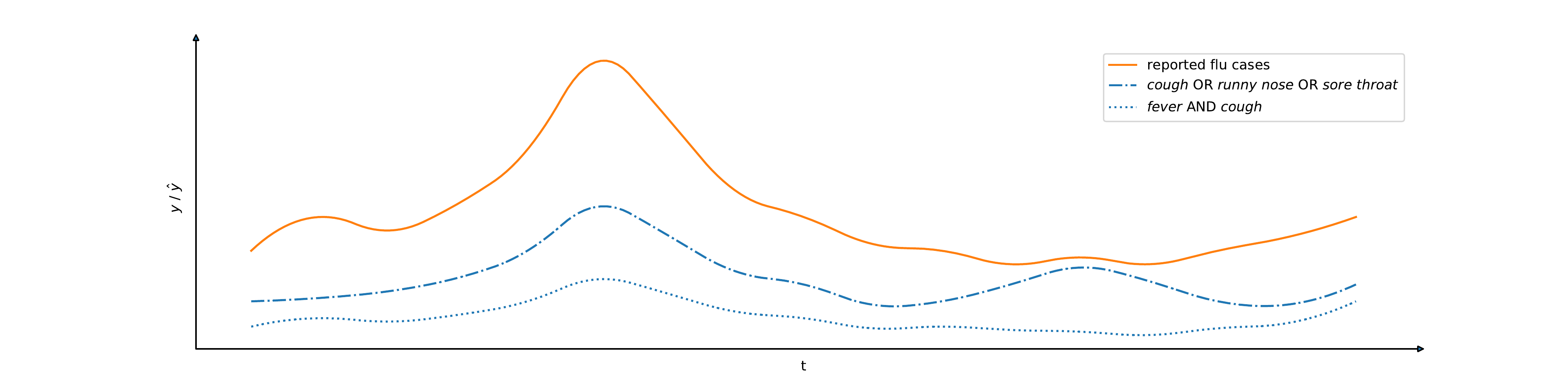}
    \caption{Exemplary comparison of two syndrome definitions (blue lines) with reported cases (orange line). The Pearson's correlation for ``{fever AND cough}'' is $0.98$ and for ``{cough OR runny nose OR sore throat}'' is $0.88$.}
    \label{fig:example}
\end{figure*}

\section{Preliminaries}

In the following, we formalize the problem that we address in the present work, including a definition of relevant notation and an overview of related work.

\subsection{Problem Definition}

\newcommand{\exspace}{\mathcal{X}}
\newcommand{\exseries}{X}
\newcommand{\exset}{X}
\renewcommand{\vec}[1]{\boldsymbol{#1}}
\newcommand{\ex}{\vec{x}}
\newcommand{\iterex}{n}
\newcommand{\numex}{N}
\newcommand{\attrset}{A}
\newcommand{\attr}{a}
\newcommand{\iterattr}{k}
\newcommand{\numattr}{K}
\newcommand{\attrval}{x}
\newcommand{\mapfun}{h}
\newcommand{\itertime}{t}
\newcommand{\numtime}{T}
\newcommand{\timespace}{\mathcal{Y}}
\newcommand{\timeseries}{\vec{y}}
\newcommand{\target}{y}
\newcommand{\classifierfun}{f}
\newcommand{\predseries}{\vec{\hat{y}}}
\newcommand{\pred}{\hat{y}}
\newcommand{\measure}{m}

We are concerned with the deduction of patterns from a health-related data source $\exseries = \left( \ex_{1}, \dots, \ex_{\numex} \right) \in \exspace$. It incorporates information about individual \emph{instances} $\ex_{\iterex} \in \exseries$ from a population $\exspace$, which are represented in terms of a finite set of predefined \emph{attributes} $\attrset = \left \{ \attr_{1}, \dots, \attr_{\numattr} \right \}$. An instance $\ex = \left( \attrval_{1}, \dots, \attrval_{\numattr} \right)$, e.g., representing a patient that has received treatment in an emergency department, assigns discrete or numerical values $\attrval_{\iterattr}$ to the $\iterattr$-th attribute $\attr_{\numattr}$. For example, discrete attributes can be used to specify a patient's gender, whereas numerical attributes are suitable to encode continuous values, such as body temperature, blood pressure or the like. The values for individual attributes may also be missing, e.g., because some medical tests have not been carried out as part of an emergency treatment. In addition, each instance in a data source is subject to a mapping $\mapfun : \mathbb{N}_{+} \rightarrow \mathbb{N}_{+}$. It associates the $\iterex$-th instance with a corresponding period in time, identified by a timestamp $\itertime = \mapfun \left( \iterex \right)$. Instances that correspond to the same interval, e.g., to the same week, are assigned the same timestamp $t: 1 \leq \itertime \leq \numtime$.

For each timestamp $t$, the instances in a data source may be associated with, a corresponding \emph{target variable} $\target_{\itertime} \in \timeseries$ to be provided as part of a secondary data source $\timeseries = \left( \target_{1}, \dots, \target_{\numtime} \right) \in \timespace$. The target space $\timespace$ corresponds to the number of infections that may occur within consecutive periods of time. Consequently, a particular target variable $\target_{\itertime} \in \mathbb{N}_{+}$ specifies how many cases related to a particular infectious disease have been reported for the $\itertime$-th time interval.


The learning task, which we address in this work, requires to find an interpretable model $\classifierfun : \exspace \rightarrow \timespace$. Given a set of instances $\exset \subset \exspace$ that are mapped to corresponding time intervals via a function $\mapfun$, it provides an estimate $\predseries = \classifierfun \left( \exset, \mapfun \right) = \left( \pred_{1}, \dots, \pred_{\numtime} \right) \in \timespace$ of the number of infections per time interval. The selection of instances and the number of reported cases, which are provided for the training of such model, must neither originate from the same source, nor comprise information about identical subgroups of the population. As a consequence, the estimates of a model are not obliged to reflect the provided target variables in terms of their absolute values. Instead, we are interested in capturing the correlation between indicators that may be derived from the training instances and the number of infections that have arised during the considered timespan. To assess the quality of a model, we compare the estimates it provides to the target variables with respect to a suitable correlation coefficient, such as  \emph{Spearman's} $\rho$, \emph{Kendall's} $\tau$, or  \emph{Pearson's} correlation. For example, one could align patient data of an medical office with local reported flu cases. Figure~\ref{fig:example} exemplary visualizes two syndrome definitions which are obtained by counting the number of patients per timestamp which fulfill a particular clinical picture. The syndrome ``{fever AND cough}'' covers less cases but it has an higher Pearson's correlation coefficient ($0.98$ compared to $0.88$).

\subsection{Related Work}

Disease patterns for syndromic surveillance are usually defined according to the knowledge of domain experts. This requires a manual examination of the available health-related data to identify indicators that may be related to a particular disease at hand. For example, \citet{edge2006} and \citet{muchaal2015} analyze information about the sales of pharmaceuticals to reason about the spread of Norovirus infections, based on their effectiveness against gastrointestinal symptoms. Similarly, the data that is gathered in emergency departments may also serve as a basis for the definition of disease patterns. In this case, definitions are usually based on the symptoms of individual patients and the diagnoses made by the medical staff. For example, \citet{ivanov2002} and \citet{suyama2003} rely on standardized codes for the \emph{International Classification of Diseases} (ICD)~\citep{trott1977}. 
\citet{boender2021syndromOnED} additionally use chief complaints of the patients at the emergency departments. 
The majority of syndrome definitions are targeted at common infectious diseases, such as gastrointestinal infections, influenza-like illnesses or respiratory diseases (e.g.,~\citep{heffernan2004, bouchouar2021, suyama2003,boender2021syndromOnED}). However, they are also used to detect other health-related epidemics, e.g., increased usage of psychoactive substances~\citep{nolan2017}.


The deduction of indicators from unstructured data, such as textual reports of complaints or diagnoses, is particularly challenging. To be able to deal with such data, text documents are often represented in terms of keywords they consist of. For example, \citet{lall2017} use syndromes that apply to the keywords contained in medical reports. Similarly, \citet{heffernan2004} use a list of exclusive keywords to reduce the chance of misclassifications, \citet{bouchouar2021} utilizes regular expressions to extract symptoms from texts and \citet{ivanov2002} use a classifier system that takes textual data as an input to assign syndromes to individual patients. In order to train a classifier, the latter approach requires labeled training data that must manually be created by experts. The analysis of textual data is even more profound in approaches to syndromic surveillance that are based on web data. For example, \citet{verlardi2014} analyze Twitter messages to capture indicators for the spread of influenza-like illnesses. Starting with a handcrafted set of medical conditions that are related to the respective disease, they learn a language model that aims to identify closely related terms based on clustering. 


The problem of learning syndrome definitions in a data-driven way, without relying on expert knowledge, has for example been addressed by \citet{kalimeri2019}. The authors of this work propose an unsupervised, probabilistic framework based on matrix factorization. Their goal is to identify patterns of symptoms in structured data that has been obtained from participatory systems. Given a set of 19 symptoms, e.g., fever or vomiting, they construct a matrix that incorporates information about the occurrences of individual symptoms over time. Ultimately, syndromes can be generated from this matrix by extracting latent features that correspond to linear combinations of groups of symptoms.

Another method that relies on structured data is proposed by \citet{goldstein2011}. It is aimed at capturing the likelihood of syndromes for a particular infectious disease. The authors propose to use expectation maximization and deconvolution to identify syndromes, which are highly correlated with the occurrences of symptoms that have been reported in regular time intervals. However, their approach does only allow to evaluate and compare disease patterns that have been specified in advance. Even though the aforementioned algorithms deal with structured data that is less cumbersome to handle than unstructured inputs, they have only be applied to small and pre-selected sets of features.


The problem of learning from assignments of target variables to sets of instances, rather than individual instances, is known as \emph{multiple instance learning} \citep{carbonneau18survey}. \citet{chevaleyreZ01MILripper} tackle such task by adapting the quality criterion used by the well-known rule learning method RIPPER. To be able to deduce classification rules from sets of instances, \citet{bjerring2011MITI} incorporate the separate-and-conquer rule induction technique into a tree learner. Both approaches are limited to the assignment of a binary signal to a bag of instances and are not intended to cope with multiple instance regression tasks \citep{ray01MILregression}. The mapping of numeric values to bags of instances, as in the syndrome definition learning task at hand, is a much less explored problem in the literature. We are not aware of any existing work that approaches this kind of problem with the goal to obtain rule-based models.

%% file: sections/3_syndrome_learning.tex
\section{Learning of Syndrome Definitions}

\newcommand{\rulmodel}{r}
\newcommand{\rul}{r}
\newcommand{\iterrul}{l}
\newcommand{\numrul}{L}
\newcommand{\cond}{c}
\newcommand{\itercond}{m}
\newcommand{\numcond}{M}
\newcommand{\minsup}{s}
\newcommand{\pearson}{\measure_{\textit{P}}}

In the following, we propose an algorithm for the automatic induction of syndrome definitions, based on the indicators that can be constructed from a health-related data source. Each indicator $\cond_{\itercond}$, which is included in such a model, refers to a certain attribute that is present in the data. It compares the values, which individual instances assign to this particular attribute, to a constant using relational operators, such as $=$ if the attribute is discrete, or $\leq$ and $>$ if it is numerical. By definition, if an indicator is concerned with an attribute for which an instance's value is missing, the indicator is not satisfied. We strive for a combination of different indicators via logical \textsc{and} ($\land$) and $\textsc{or}$ ($\lor$) operators. The model that is eventually produced is given in \emph{disjunctive normal form}, i.e., as a disjunction of conjunctions. 
Such a logical expression $\rulmodel = \rul_{1} \lor \dots \lor \rul_{\numrul}$ with $\rul_{\iterrul} = \cond_{\iterrul, 1} \land \dots \land \cond_{\iterrul, \numcond}$ evaluates to $\rulmodel \left( \ex_{\iterex} \right) = 1$ (\emph{true}) or $\rulmodel \left( \ex_{\iterex} \right) = 0$ (\emph{false}), depending on whether it is satisfied by a given instance $\ex_{\iterex}$ or not. If the context is clear, we abbreviate $\cond_{\iterrul,i}$ with $\cond_{i}$. The number of infected cases, which are estimated by a logical expression $\rulmodel$ for individual time intervals $\itertime$, calculate as
\begin{equation*}
\predseries = \rulmodel \left( \exset \right) = \left( \sum \nolimits_{\ex_{\iterex} \in \exset} \llbracket \mapfun\left( \iterex \right) = \itertime \rrbracket  \rulmodel \left( \ex_{\iterex} \right) \right)_{1 \leq \itertime \leq \numtime},
\end{equation*}
where $\llbracket p \rrbracket = 1$ if the predicate $p$ is true, and $0$ otherwise.


The representation of syndromes introduced above is closely related to sets of conjunctive rules $\rul_{\iterrul}$ as commonly used in \emph{inductive rule learning} --- an established and well-researched area of machine learning (see, e.g., \citep{fuernkranz2012} for an overview on the topic). Consequently, we rely on commonly used techniques from this particular field of research to learn the definitions of syndromes. We use a sequential algorithm that starts with an empty hypothesis to which new conjunctions of indicators $\rul_{1}, \dots, \rul_{\numrul}$ are added step by step. Given a data source that incorporates many features, the number of possible combinations of indicators can be very large. For this reason, we rely on \emph{top-down hill climbing} to search for suitable combinations. With such an approach, conjunctions of indicators that can potentially be added to a model are constructed greedily. At first, single indicators are taken into account individually. They are evaluated relative to the existing model and the one that promises the highest improvement in quality is ultimately selected. Afterwards, it is iteratively refined by evaluating the combinations that possibly result from a conjunction of already chosen indicators with an additional one. The search continues to add more indicators, resulting in more restricted patterns that apply to fewer instances, as long as an improvement of the model's quality can be achieved. Optionally, the maximum number of indicators per conjunction $\numcond$ can be limited via a parameter. If $\numcond = 1$, the algorithm is restricted to learn disjunctions of indicators. Furthermore, we enforce a \emph{minimum support} $\minsup \in \mathbb{R}$ with $0 < \minsup < 1$, which specifies the number of instances $\numex \cdot \minsup$ a conjunction of indicators must apply to. Once it has decided for a conjunction of indicators to be included in the model, the algorithm attempts to learn another conjunction to deal with instances that have not yet been adequately addressed by the model. The training procedure terminates as soon as it is unable to find a new pattern that improves upon the quality of the model. In addition, an upper bound can be imposed on the number of disjunctions $\numrul$ by the user.


The search for suitable indicators and combinations thereof is guided by a target function to be optimized at each training iteration. It assesses the quality that results from adding an additional conjunction of indicators to an existing model in terms of a numeric score. We denote the estimates that are provided by a model after the $\iterrul$-th iteration as $\predseries^{\left( \iterrul \right)}$. When adding a conjunction of indicators $\rul_{\iterrul}$ to an existing model, the estimates of the modified model can be computed incrementally as
\begin{equation*}
\predseries^{\left( \iterrul \right)} = \rulmodel^{\left( \iterrul \right)} \left( \exset \right) = \rulmodel^{\left( \iterrul - 1 \right) } \left( \exset \right) + \rul_{\iterrul} \left( \exset \right).
\end{equation*}
We assess the quality of a model's estimates in terms of the absolute \emph{Pearson correlation coefficient}. It can be computed in a single pass over the target time series $\timeseries$ and the corresponding estimates $\predseries = \classifierfun\left( \exseries \right)$ according to the formula
\[
\pearson \left( \timeseries, \predseries^{\left( \iterrul \right)} \right) \coloneqq \left| \frac{\numtime \sum \target_{\itertime} \pred_{\itertime} \sum \target_{\itertime} \sum \pred_{\itertime}}{\sqrt{\numtime \sum \target_{\itertime}^2 - \left( \sum \target_{\itertime} \right)^2} \sqrt{\numtime \sum \pred_{\itertime}^2 - \left( \sum \pred_{\itertime} \right)^2}} \right|.
\]
If the score that is computed for a potential modification according to the target function $\pearson$ is greater than the quality of the current model, it is considered an improvement. Among all possible modifications that are considered during a particular training iteration, the one with the greatest score is preferred.

%% file: sections/4_evaluation.tex
\section{Evaluation}

To evaluate the previously proposed learning approach, we have implemented the methodology introduced above by making use of the publicly available source code of the BOOMER rule learning algorithm~\citep{rapp2020}. In adherance to the principles of reproducible research, our implementation can be accessed online\footnote{\url{https://github.com/mrapp-ke/SyndromeLearner}}. A major goal of the empirical study, which is outlined in the following, is to investigate whether the proposed methodology is able to deduce patterns from health-related data that correlate with the number of infections supplied via a secondary data source. For our experiments, we relied on routinely collected and fully anonymized data from 12 German emergency departments which capture information about patients that have consulted these institutions between January 2017 and April 2021. 

In a first step, we conducted a series of experiments using synthetic syndrome definitions. 
The objective was to validate the algorithm and to better understand its capabilities and limitations when it comes to the reconstruction of known disease patterns in a controlled environment. On the one hand, we considered synthetic syndromes with varying characteristics and complexity. On the other hand, we investigated the impact that the temporal granularity of the available data has on the learning approach. As elaborated below, the health-related data used in this work are available on a daily basis. By using synthetic syndromes, we were able to validate the algorithm's behavior when dealing with a broader or more fine-grained granularity as well. 
The use of synthetic syndromes also allows to investigate the ability of the proposed approach  independently of the negative effects of artifacts that may be present in real data. 
This includes delays of reports, inaccuracies in the reported dates or instances that are present in one data source, but not in the other. For example, cases may have been reported in one of the considered districts, but have not been treated in one of the emergency departments included in our dataset. Vice versa, it is also possible that cases have been treated at one of the considered departments but have not been reported to the public agencies.


Such artifacts almost certainly play a role in our second experiment, where  we tried to discover patterns that correlate with the publicly reported cases.
We selected cases from the notifiable diseases
of \emph{Influenza} and \emph{Norovirus}, which have extensively been studied in existing work (e.g.,~\citep{heffernan2004,muchaal2015,kalimeri2019}), as well as of the recently emerged \emph{SARS-CoV-2}, which has for example been analyzed by \citet{bouchouar2021}. 
To evaluate whether the algorithm is able to identify meaningful indicators that are related to these particular diseases, we provide a detailed discussion of the discovered syndromes and compare them to manually defined disease patterns.


\subsection{Experimental Setup}

\subsubsection{Health-related Data}

\input{sections/ed_data_table}

As shown in Table~\ref{tab:emergency_department_attributes}, we have extracted 15 attributes from the emergency department data. Each of the available attributes corresponds to one out of four categories. The first category, \emph{diagnosis}, includes an initial assessment in terms of the \emph{Manchester Triage System} (MTS)~\citep{graeff2014}. It is obtained for each patient upon arrival at an emergency department. Besides, this first category also comprises an ICD~\citep{trott1977} code that represents a physician's assessment. In addition to the full ICD code, we also consider a more general variant that consists of the leading character and the first two digits (e.g., \textit{U07} instead of \textit{U07.1}). Features that belong to second category, \emph{demographic information}, indicate the gender and age of patients, whereas \emph{vital parameters} correspond to measurement data, such as blood pressure or pulse frequency, that may have been registered by medical staff. Features of the last category, \emph{contextual information}, may provide information about why a patient was possibly quarantined (\emph{isolation}), the means of transport used to get to the emergency department (\emph{transport}) and the status when exiting the department (\emph{disposition}).

In contrast to existing work on the detection of disease patterns (e.g.,~\citep{goldstein2011,kalimeri2019}), we have not applied any pre-processing techniques to the health-related data, such as a manual selection of symptoms that are known to be related to an infectious disease. As a consequence, the data contains a lot of noise, e.g., diagnoses related to injuries, and many missing values (cf.~Table~\ref{tab:emergency_department_attributes}). In accordance with the findings of \citet{hartnett2020}, we observed a reduced number of emergency department visits during the first weeks of the SARS-CoV-2 pandemic. However, preliminary experiments suggested that this anomaly has no effect on the operation of our algorithm. To obtain a single dataset, we have merged the data from the considered emergency departments. It consists of approximately 1,900,000 instances. Each of the instances corresponds to a particular week (i.e., around 8,500 instances per week). Additional information about the emergency data used in this work is provided by \citet{boender2021syndromOnED}, who used a slightly different subset of the data set to evaluate their handcrafted syndrome definitions.

\subsubsection{Number of Infections}

The number of cases corresponding to the infectious diseases Influenza, Norovirus and SARS-CoV-2 have been retrieved from the \emph{SurvStat}\footnote{\url{https://survstat.rki.de}} platform provided by the \emph{Robert Koch-Institut}. To match the temporal information in the health-related dataset, we have aggregated the weekly reported numbers for German districts (``Landkreise'' and ``Stadtkreise'') where the considered emergency departments are located.

\subsubsection{Parameter Setting}

For all experiments that are discussed in the following, we have set the minimum support to $\minsup = 0.0001$. With respect to the approximately 1,900,000 instances contained in the training dataset, this means that each conjunction of indicators considered by the algorithm must apply to at least $190$ patients. In preliminary experiments we have found this setting to produce reasonable results, while keeping the training time at an acceptable level (typically under one minute). In addition, we have limited the maximum number of disjunctions in a model to $\numrul = 50$. However, the algorithm usually terminates before this number is reached.

\subsection{Reconstruction of Synthetic Syndromes}

\begin{figure}[t]
    \centering
    \includegraphics[width=0.9\columnwidth]{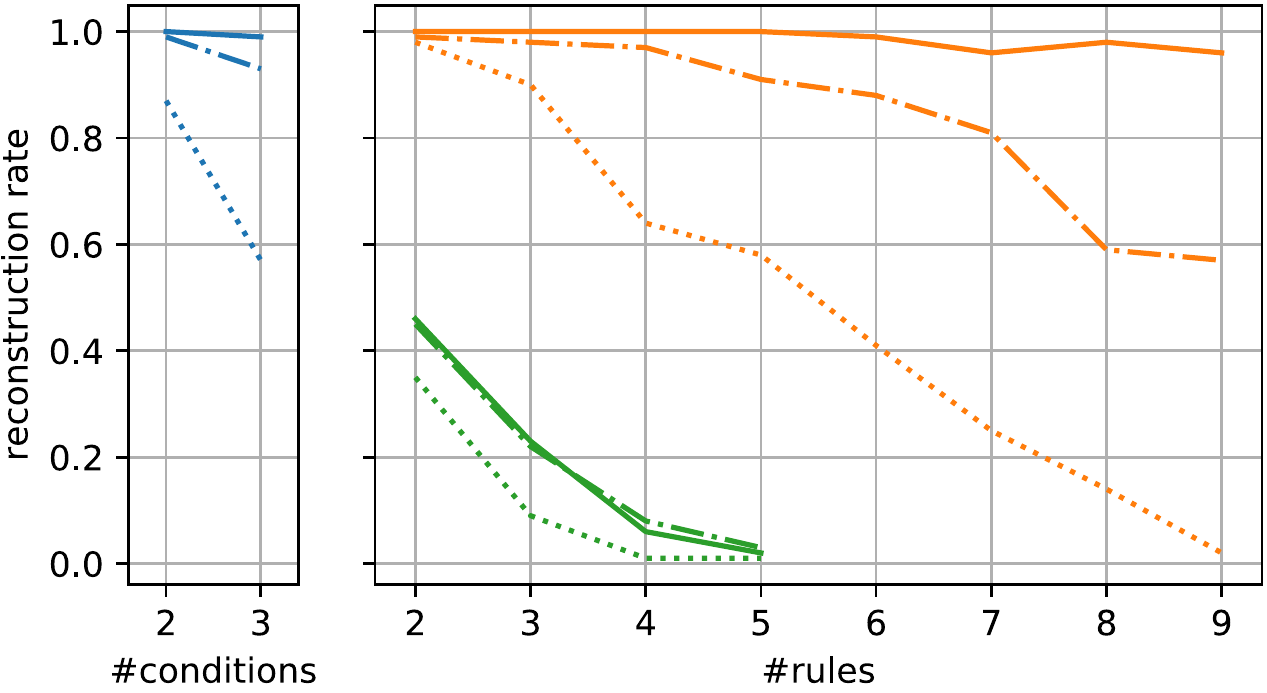}\\
    \vspace{0.2cm}
    \includegraphics[width=0.9\columnwidth]{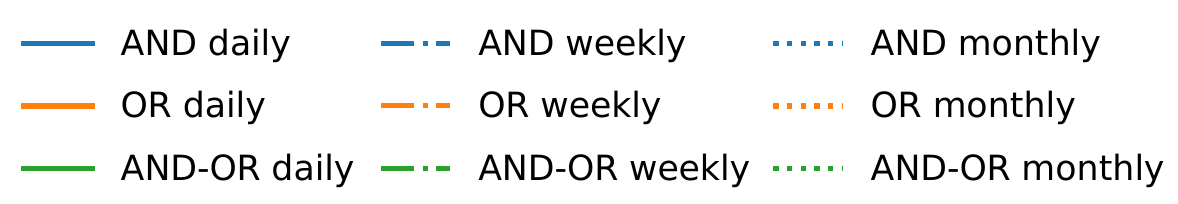}
    \caption{Percentage of successfully reconstructed syndrome definitions of different types for varying complexities of the  predefined syndromes.} 
    \label{fig:synthetic_evaluation}
\end{figure}

In our first experiment, we validated the ability of our algorithm to discover disease patterns under the assumption that the reported cases are actually present in the data. For this purpose, we defined synthetic syndromes with varying characteristics from the emergency department data. For each syndrome, we determined the number of instances they apply to over time. The goal of the algorithm was to reconstruct the original syndrome definitions, exclusively based on the correlation with the corresponding number of cases. For this experiment, we focused on syndromes that use ICD codes and MTS representations, since these indicators are most commonly used in related work (e.g., \citep{suyama2003,ivanov2002,boender2021syndromOnED}). We have not used short versions of the ICD codes due to their overlap with the full codes. 
The following three different types of synthetic syndromes were considered:
\begin{enumerate}
    \item Conjunctions of indicators (\textsc{and}): \[ \rul = \cond_{1} \land \ldots \land \cond_{\numcond},\ \text{ where}\ \numcond \in \left \{ 2,3 \right \} \]
    \item Disjunctions of indicators (\textsc{or}): \[ \rul_{1} \lor \ldots \lor \rul_{\numrul},\ \text{where}\ \rul_{\iterrul} = \cond\ \text{and}\ \numrul \in \left[ 2, 9\right] \]
    \item Disjunctions of conjunctions (\textsc{and-or}): \[ \rul_{1} \lor \ldots \lor \rul_{\numrul},\ \text{where}\ \rul_{\iterrul} = \cond_{1} \land \cond_{2}\ \text{and}\ \numrul \in \left[ 2, 5 \right] \]
\end{enumerate}
For each syndrome type, we generated 100 artificial definitions by randomly selecting indicators that are present in the data, such that each indicator and each conjunction of indicators applies to at least 200 patients. This ensures that the syndromes that are ultimately generated apply to this particular number of patients at minimum. In addition, we have considered three temporal granularities to determine the number of cases different syndromes apply to. Experiments have been conducted with counts that are available on a daily, weekly or monthly basis. To quantify to which extent our approach is able to reconstruct the original syndrome definitions, we compute the percentage of correctly identified patterns, i.e., syndromes that use the exact same indicators, referred to as the \emph{reconstruction rate}. A visualization of the experimental results is given in Figure~\ref{fig:synthetic_evaluation}.

Generally, we can observe that the algorithm's ability to capture the predefined disease patterns benefits from a more fine-grained granularity of the available data (e.g., daily instead of weekly reported numbers). This meets our expectations, as a greater temporal resolution results in  more specific patterns of covered cases, given a particular syndrome. As a result, it is easier to identify the indicators that allow to replicate a certain disease pattern and separate them from unrelated ones. In particular, syndromes that are exclusively based on disjunctions (\textsc{or}) or conjunctions (\textsc{and}), regardless of their complexity, can reliably be captured when supplied with daily numbers. When dealing with a broader temporal granularity, the uniqueness of disease patterns vanishes and they become more likely to interfere with the numbers resulting from similar syndromes.


Regarding the different types of predefined syndromes, it can be seen that their reconstruction becomes more difficult as their complexity increases. Especially when dealing with syndromes that include both, disjunctions and conjunctions (\textsc{and-or}), the reconstruction rate mostly depends on the number of indicators, whereas the temporal resolution plays a less important role. This shows the limitations of a greedy hill climbing strategy when it comes to the reconstruction of complex patterns. To overcome these shortcomings, approaches for the re-examination of previously induced rules, such as pruning techniques, could be considered. It is also possible to extend the search space that is explored by the training algorithm, e.g., by conducting a beam search, where several promising solutions are explored instead of focusing on a single one at each step. However, if the patterns, which have been found by the algorithm, only slightly differ from the predefined syndromes (e.g., by omitting or including infrequent ICD codes). While we did not evaluate this in depth, we believe they could still 
comprise useful information, e.g., by providing alternative, but nearly equivalent, descriptions of the syndrome.

\begin{figure*}[t]
\centering

\subfloat[Influenza\label{fig:influenza}]{%
\begin{tabular}{c}
\includegraphics[width=0.3\textwidth]{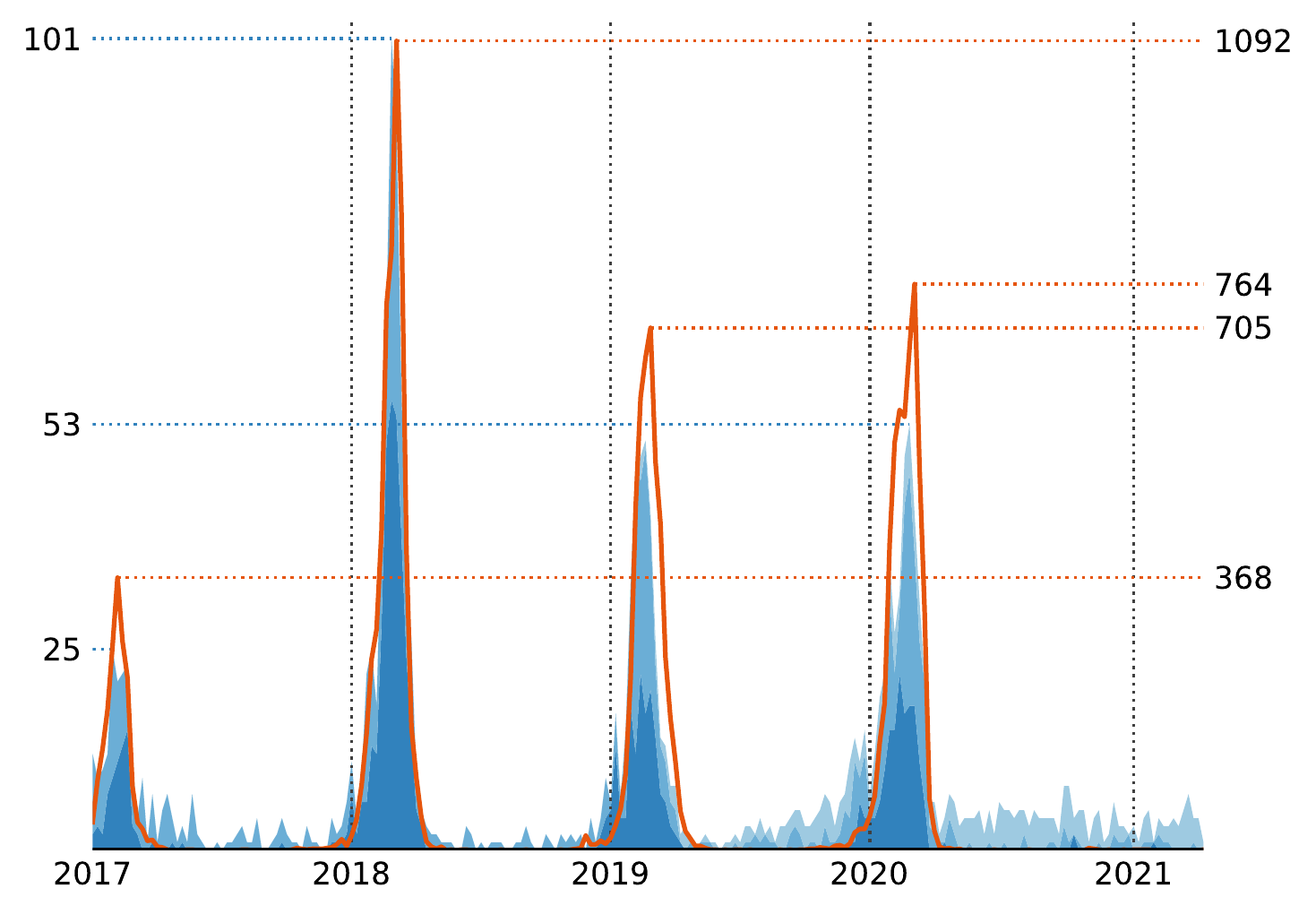} 
\\
\includegraphics[width=0.3\textwidth]{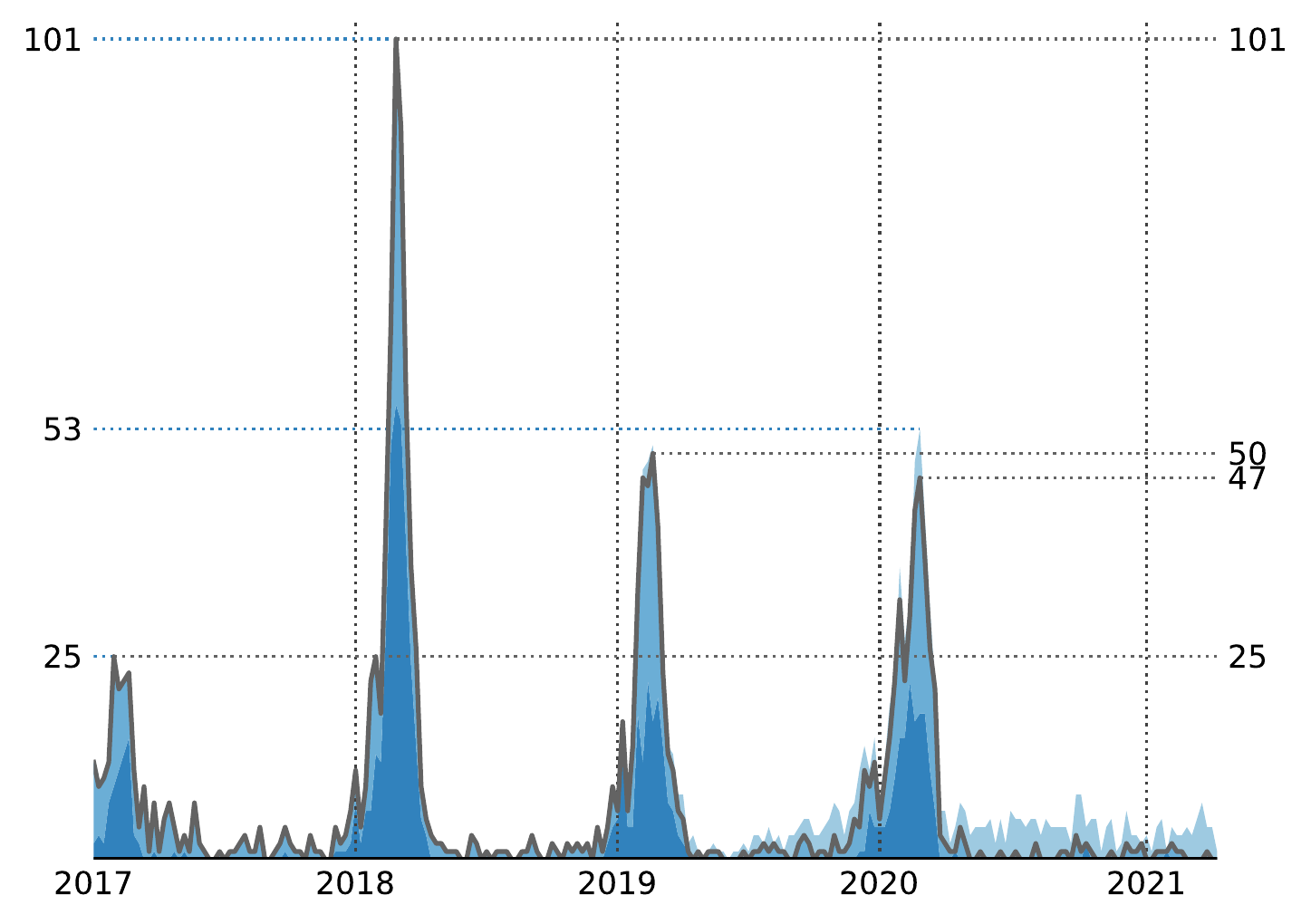} 
\end{tabular}
}
%
\subfloat[SARS-CoV-2\label{fig:covid}]{\begin{tabular}{c}
\includegraphics[width=0.3\textwidth]{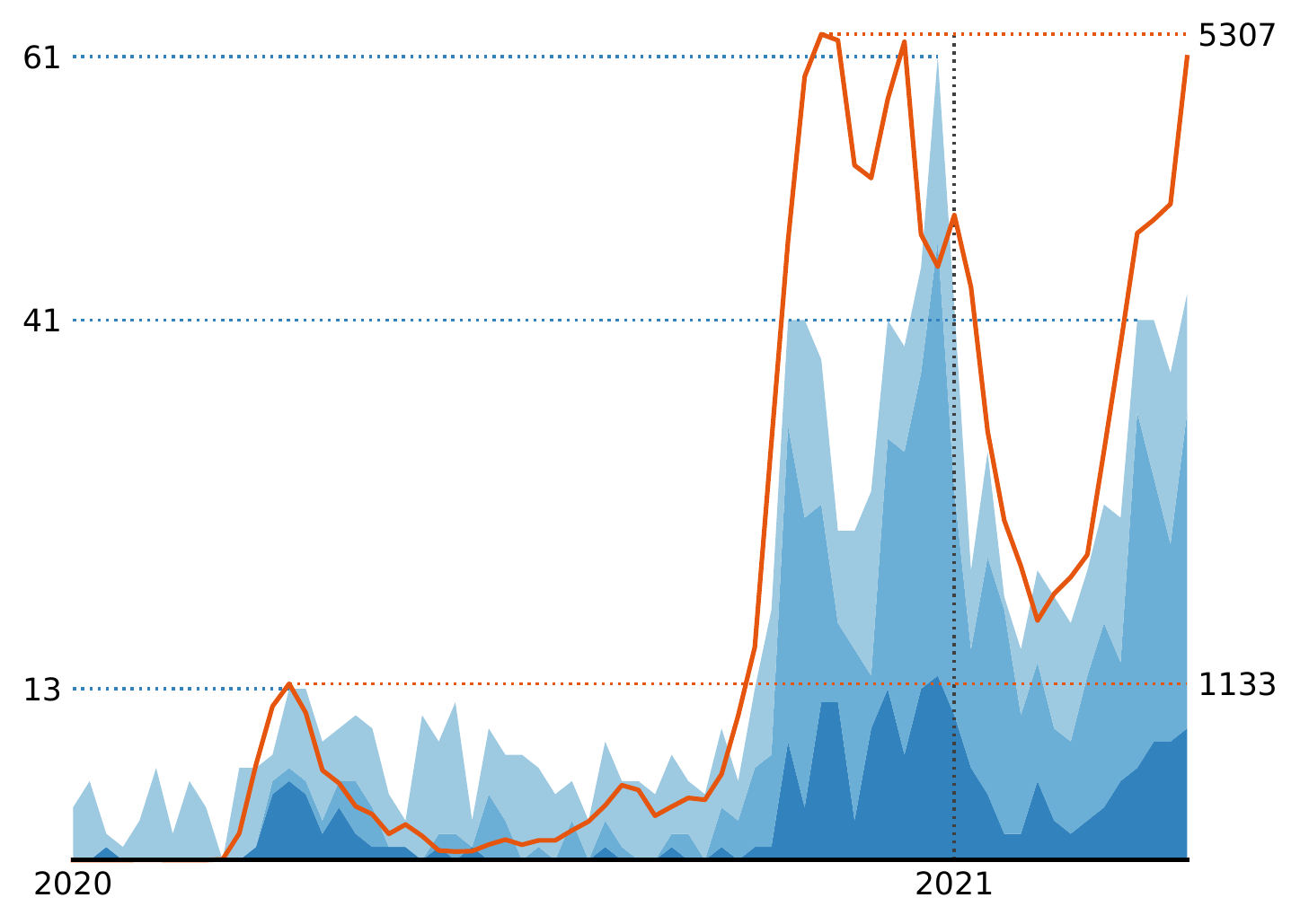} 
\\
\includegraphics[width=0.3\textwidth]{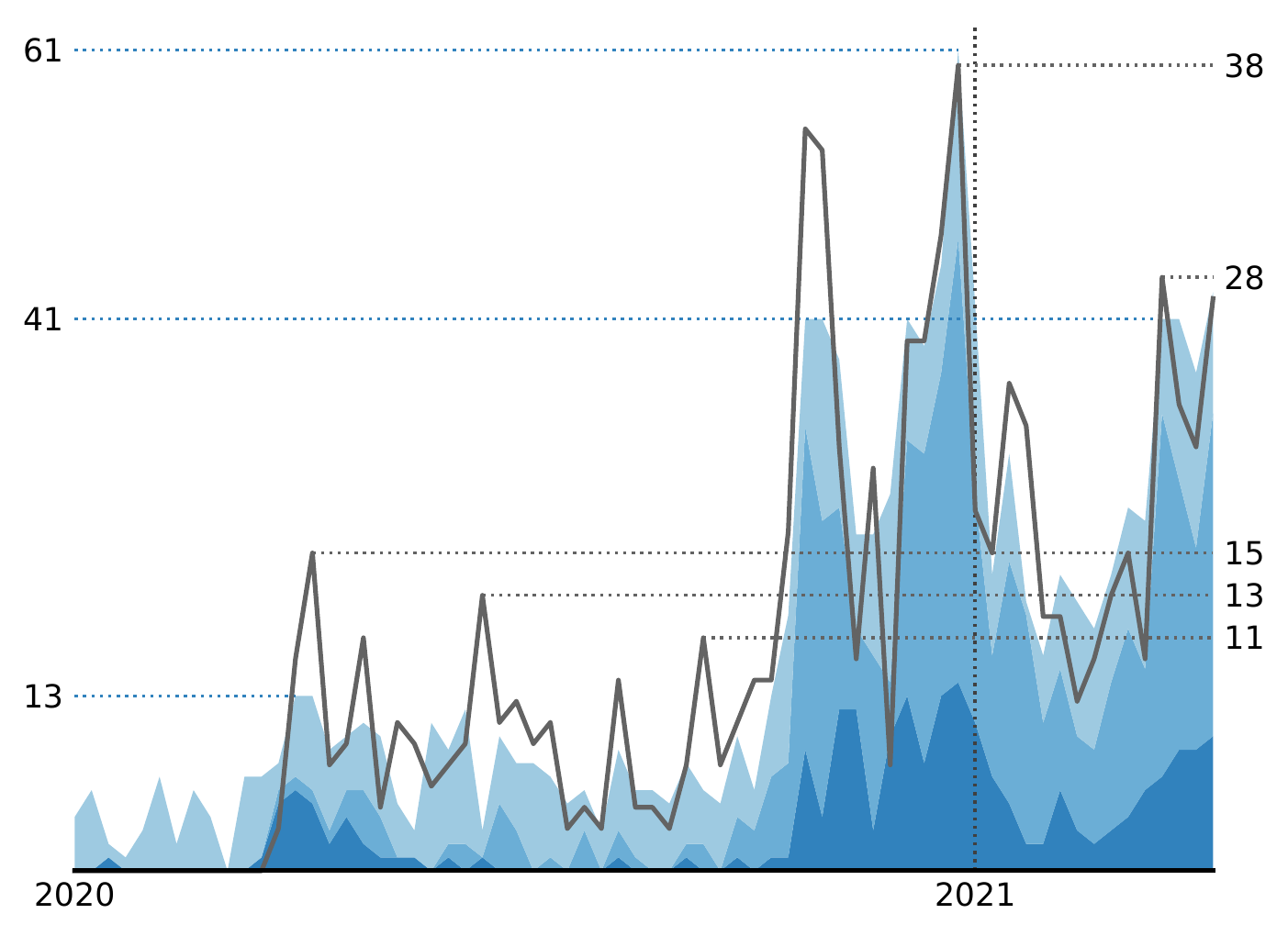} 
\end{tabular}
}
%
\subfloat[Norovirus\label{fig:norovirus}]{%
\begin{tabular}{c}
\includegraphics[width=0.3\textwidth]{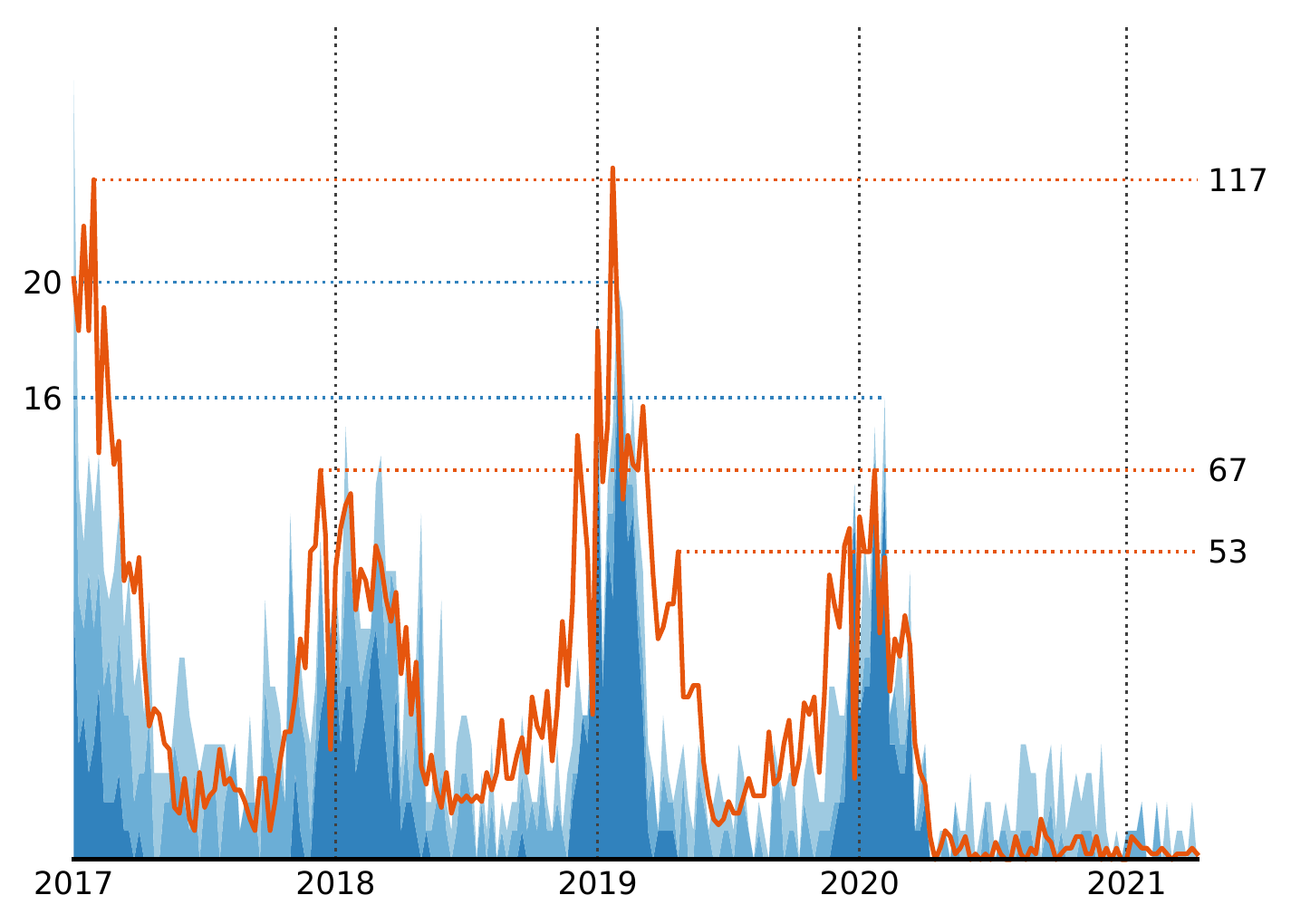} 
\\
\includegraphics[width=0.3\textwidth]{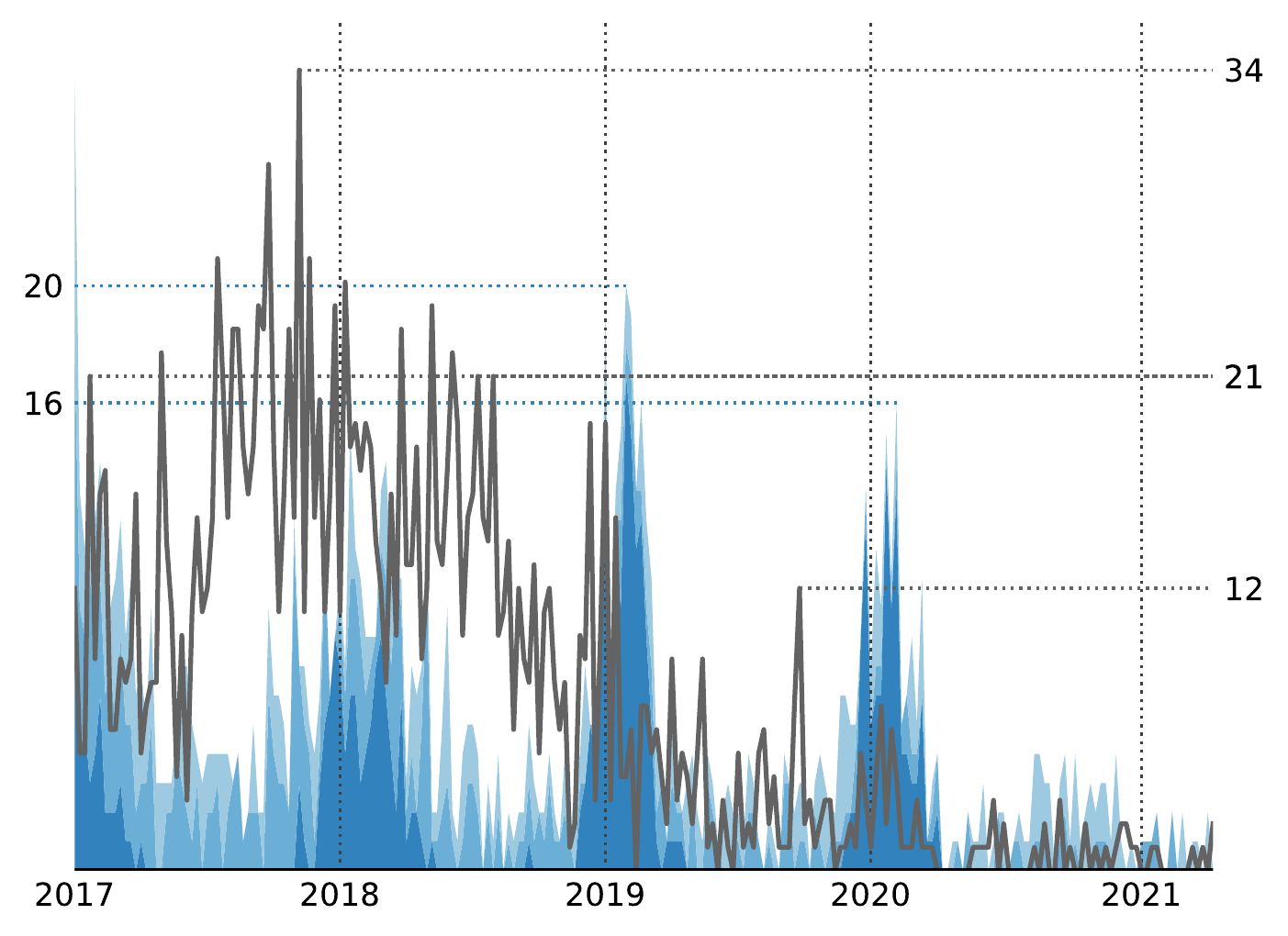} 

\end{tabular}
}
\caption{Number of cases satisfying the discovered syndrome definitions compared to actual cases (top, orange) and handcrafted syndromes (bottom, black).}\label{fig:comparison}
\end{figure*}

\subsection{Discovery of Syndrome Definitions from Real-World Data}

\begin{table}[t]
\caption{Pearson correlation between cases identified by automatically learned syndromes on different feature categories and actually reported cases as well as cases that match the handcrafted syndrome definitions.}
\label{tab:pearson}
\centering
\begin{tabular}{c c c c c r r}
\toprule
\hspace{0.1cm} & \multicolumn{4}{c}{\textbf{feature categories used}} & \multicolumn{1}{c}{\textbf{reported}} & \multicolumn{1}{c}{\textbf{handcrafted}} \\
& \textbf{\Circled{1}} & \textbf{\Circled{2}} & \textbf{\Circled{3}} & \textbf{\Circled{4}} & \multicolumn{1}{c}{\textbf{cases}} & \multicolumn{1}{c}{\textbf{syndromes}} \\
\midrule
\multicolumn{7}{l}{\textbf{Influenza}} \\
& \checkmark &            &            &            & 0.9354 & 0.9917 \\
& \checkmark & \checkmark &            &            & 0.9357 & 0.9796 \\
& \checkmark &            & \checkmark &            & 0.9480 & 0.9768 \\
& \checkmark &            &            & \checkmark & 0.9366 & 0.9948 \\
& \checkmark & \checkmark & \checkmark & \checkmark & 0.9493 & 0.9800 \\
\midrule
\multicolumn{7}{l}{\textbf{SARS-CoV-2}} \\
& \checkmark &            &            &            & 0.9399 & 0.9473 \\
& \checkmark & \checkmark &            &            & 0.9454 & 0.9219 \\
& \checkmark &            & \checkmark &            & 0.9528 & 0.8689 \\
& \checkmark &            &            & \checkmark & 0.9464 & 0.9506 \\
& \checkmark & \checkmark & \checkmark & \checkmark & 0.9528 & 0.8689 \\
\midrule
\multicolumn{7}{l}{\textbf{Norovirus}} \\
& \checkmark &            &            &            & 0.7669 & 0.2761 \\
& \checkmark & \checkmark &            &            & 0.7669 & 0.2761 \\
& \checkmark &            & \checkmark &            & 0.7303 & 0.1470 \\
& \checkmark &            &            & \checkmark & 0.7167 & 0.1608 \\
& \checkmark & \checkmark & \checkmark & \checkmark & 0.7242 & 0.1672 \\
\bottomrule
\end{tabular}
\end{table}

In our second experiment, we used the proposed algorithm to obtain syndrome definitions for the infectious diseases Influenza, Norovirus and SARS-CoV-2. In the literature, the quality of syndromes is either evaluated by experts (e.g., \citep{ivanov2002,heffernan2004,lall2017,bouchouar2021}) or by measuring the correlation with reported infections, reported deaths or expert definitions (e.g., \citep{nolan2017,verlardi2014,kalimeri2019,suyama2003,muchaal2015,edge2006}). We follow the latter approach by reporting the Pearson correlation coefficient of the automatically discovered disease patterns with the publicly reported number of infections supplied for training, as well as syndromes that have been handcrafted by ourselves. In addition, we provide a detailed discussion of the indicators included in our models.

Inspired by the expert syndrome definitions for Influenza and SARS-CoV-2 used by \citet{boender2021syndromOnED}, we created a set of similar, but much simpler, definitions solely based on ICD codes. They incorporate the ICD codes that correspond to suspected or confirmed cases of a particular disease, i.e., \textit{J10} (Influenza due to identified seasonal influenza virus) or \textit{J11} (Influenza, virus not identified) for Influenza, \textit{A08} (viral and other specified intestinal infections) for Norovirus and \textit{U07.1} (COVID-19, virus identified) or \textit{U07.2} (COVID-19, virus not identified) for SARS-CoV-2. We have found the number of cases these ICD codes apply to be very similar to those matched by the aforementioned expert definitions.

For each of the considered diseases, we trained several models using different sets of features. First of all, for a fair comparison with the handcrafted syndromes, we provided our algorithm with the features that belong to the first category in Table~\ref{tab:emergency_department_attributes}, i.e., ICD codes and MTS representations. A visualization of the number of infections that correspond to the disease patterns that have been discovered with respect to these features is shown in Figure~\ref{fig:comparison}. Each one of them includes a comparison with the reported number of infections supplied for training and the number of cases our handcrafted syndromes apply to, respectively. In the case of Influenza and SARS-CoV-2, all of these numbers are strongly correlated. In the first case, one can clearly observe an increase of infections during the first months of each year. In the latter case, the different peaks of SARS-CoV-2 infections according to the publicly reported numbers are replicated by both, the handcrafted syndromes and the automatically learned patterns. The correlation between syndromes and reported numbers is less strong with respect to Norovirus. However, compared to the handcrafted syndromes, the automatically discovered patterns appear to better capture the seasonal outbreaks of this particular disease. In general, the numbers that correspond to the syndrome definitions are much lower than the reported numbers, as only a small fraction of detected cases have actually been treated in emergency departments.

In addition to ICD codes and MTS representations, we have also conducted experiments, where we provided the algorithm with one additional set of features, as well as with all features available. To validate whether the availability of additional features comes with an advantage for an accurate reproduction of the infected cases, we rely on the Pearson correlation coefficients that result from different feature selections in Table~\ref{tab:pearson}. For all experiments, we report the correlation of the autonomously learned syndromes with both, the number of reported cases used for training and the cases captured by the handcrafted syndromes. In the case of Influenza and SARS-CoV-2, the inclusion of vital parameters introduces a minor advantage for matching the reported numbers. Understandably, the use of additional features typically reduces correlation with the handcrafted syndromes, as they do not make use of these features. In the case of Norovirus, the Pearson correlation does not benefit from the availability of vital parameters. Regardless of any specific disease, this does also apply to the contextual and demographic information. We consider the absence of demographic indicators as positive, as none of the diseases appears to be specific to gender or age.

\subsection{Discussion of Discovered Syndrome Definitions}

As the use of ICD codes and MTS representations is sufficient in most cases to match the reported number of infections, we mostly focus on models that have been trained with respect to these features in the following discussion. A selection of exemplary syndromes that have been learned by our algorithm is also shown in Table~\ref{tab:syndromes}.

\subsubsection{Influenza}

The indicators that have been selected by our algorithm for modeling the number of Influenza cases include the ICD codes \textit{J10} and \textit{J11} that are also included in our handcrafted definition. These indicators have been selected during the first iterations of the algorithm and therefore are considered more important than the subsequent ones. As indicated by using different shades of blue in Figure~\ref{fig:influenza}, patterns found during early iterations (dark blue) mostly focus on the strongly pronounced seasonal peaks. Indicators that have been selected at later iterations (lighter blue) are more likely to match irrelevant cases and hence are often unrelated to the respective disease. In the case of Influenza, this includes clearly irrelevant ICD codes, such as \textit{Z96.0} (presence of urogenital implants) or \textit{S53.1} (dislocation of elbow, unspecified) as fourth and fifth indicator, but also codes that may be related to Influenza-like illnesses, such as \textit{J18.8} (other pneumonia) or \textit{J34.2} (deviated nasal septum) at positions 10 and 15. When the algorithm has access to vital parameters, the indicator \textit{J11} is combined with information about blood pressure and body temperature as follows:
\begin{align*}
\text{J11 } \land\ & \text{blood pressure diastolic} \leq 92.5 \\
\land\ & \text{blood pressure systolic} \leq 156.5 \\
\land\ & \text{temperature} > 38.5
\end{align*}
Due to the lack of domain knowledge, we are not qualified to decide whether such a pattern is in fact characteristic of Influenza. This shows the demand for experts, who are indispensable for the evaluation of machine-learned models.

\subsubsection{SARS-CoV-2} When used to learn patterns for SARS-CoV-2, our algorithm considers the MTS presentation ``breathing problem'', as well as the ICD codes \textit{J12} (viral pneumonia) and \textit{U07.1} (COVID-19, virus identified), as most relevant. The latter of these ICD codes is also included in the handcrafted syndrome definition. Besides clearly irrelevant indicators, it further selects the ICD code \textit{J34.2} (deviated nasal septum)  at a later stage of training that may be related to this particular illness. When provided with vital parameters, the algorithm decides to use the ICD code \textit{J12} in combination with data about a patient's blood pressure and temperature in its most relevant pattern:
\begin{align*}
\text{J12 } \land\ & 81.5 < \text{blood pressure systolic}  \leq 149.5 \\
\land\ & \text{blood pressure diastolic}  \leq 77.5 \\
\land\ & \text{temperature} > 36.5 
\end{align*}

\subsubsection{Norovirus} When it comes to modeling the infections with Norovirus, the algorithm fails to identify any ICD codes that are related to this particular illness, such as the ones included in our manual definition or codes related to symptoms like diarrhea. Instead, it uses indicators like \textit{J21.0} (Acute bronchiolitis due to respiratory syncytial virus) or  \textit{J34} (Other disorders of nose and nasal sinuses) in combination with other indicators to match the reported numbers. This is most probably due to the similar seasonality of Norovirus and Influenza-like illnesses. This illustrates another difficulty one may encounter when pursuing a data-driven approach to syndromic surveillance. If high numbers of infections with respect to multiple diseases occur during a similar timespan, the algorithm is not able to distinguish between indicators that relate to different types of infections. In such case it is necessary to provide additional knowledge to the learning algorithm, as it is unable to grasp the semantics of individual features on its own. In particular, this motivates the need for an interactive learning approach, where a human expert interacts with the computer in order to guide the construction of models. For example, by prohibiting the use of certain indicators or features that have been identified to be irrelevant to the problem at hand.

\begin{table}[t]
\caption{Exemplary automatically induced syndrome definitions.}
\label{tab:syndromes}
\centering
\begin{tabular}{p{0.95\columnwidth}}
\toprule
\Circled{1} \textbf{Influenza} \\
J10 $\lor$ J11 $\lor$ ``new confusion condition`` $\lor$ Z96.0 $\lor$ \ldots\\
\midrule
\Circled{1}  \textbf{SARS-CoV-2} \\
(J12 $\land$ ``breathing problems'') $\lor$ U07.1 $\lor$ ``pain in lower abdomen'' $\lor$ \ldots \\
\midrule
\Circled{1}  \textbf{Norovirus} \\
J21.0 $\lor$ D40 $\lor$ (J34 $\land$ ``recent problem``) \\
\midrule
\Circled{1}  \Circled{3} \textbf{Influenza} \\
J10 \\
 $\lor$ \textbf{(}J11 $\land$ diastolic $\leq 92.5$ $\land$ systolic $\leq 156.5$ $\land$ temperature $> 38.5$) \\
$\lor$ (temperature $ \leq 40.5$ $\land$ diastolic $\leq 108.5$ $\land$ \\
~~~~~~~~~~systolic $\leq 162$ $\land$ $187.5 \leq $ heart rate $\leq 207.5$) \\
$\lor$ \ldots \\
\midrule
\Circled{1} \Circled{2} \Circled{3} \Circled{4} \textbf{Influenza} \\
J10  \\
$\lor$ (J11 $\land$ diastolic $\leq 92.5$ $\land$ systolic $\leq 156.5$ $\land$ temperature $> 38.5$ ) \\
$\lor$ (temperature $ \leq 40.5$ $\land$ diastolic $\leq 110$ $\land$ systolic $\leq 162$ $\land$ \\
~~~~~~~$ 187.5 \leq$ heart rate $\leq 212.5$ $\land$ no isolation $\land$ patient sent home)
\\  $\lor$ 
\ldots \\
\bottomrule
\end{tabular}

\medskip 
\begin{footnotesize}
\begingroup
\setlength{\tabcolsep}{2pt} 
\begin{tabular}{rl}
\footnotesize{D40 =} & {Neoplasm of uncertain/unknown behaviour of male genital organs}\\
\footnotesize{J10 =} & {Influenza due to identified seasonal influenza virus}\\
\footnotesize{J11 =} & {Influenza, virus not identified}\\
\footnotesize{J12 =} & {Viral pneumonia, not elsewhere classified}\\
\footnotesize{J21.0 =} & {Acute bronchiolitis due to respiratory syncytial virus}\\
\footnotesize{J34 =} & {Other disorders of nose and nasal sinuses}\\
\footnotesize{U07.1 =} & {COVID-19, virus identified}\\
\footnotesize{Z96.0 =} & {Presence of urogenital implants}
\end{tabular}
\endgroup
\end{footnotesize}
\end{table}

%% file: sections/ed_data_table.tex
\begin{table}[t]
    \caption{Attributes included in the emergency department data.}
    \label{tab:emergency_department_attributes}
    \centering
    \begin{tabular}{l l r r}
        \toprule
         &  &   & \multicolumn{1}{c}{\textbf{missing}}  \\
         \hspace{0.75cm}\textbf{name} & \textbf{type} & \textbf{\#values} & \multicolumn{1}{c}{\textbf{values in \%}} \\
        \midrule
        \textbf{\Circled{1} diagnosis}\\
         \hspace{0.75cm}MTS presentation & discrete & 57 & 0.01 \\
         \hspace{0.75cm}MTS indicator & discrete & 179 & 5.10 \\
         \hspace{0.75cm}ICD code & discrete & 5901 & 65.45 \\
         \hspace{0.75cm}ICD code (short) & discrete & 1509 & 65.45\\
         \midrule
        \textbf{\Circled{2} demographic information}\\
         \hspace{0.75cm}gender &  discrete & 3 & 0.00 \\
         \hspace{0.75cm}age & discrete & 21 & 0.00 \\ 
         \midrule
         \textbf{\Circled{3} vital parameters}\\
         \hspace{0.75cm}blood pressure systolic & numeric & - & 57.19\\
         \hspace{0.75cm}blood pressure diastolic & numeric & - & 57.22\\
         \hspace{0.75cm}temperature & numeric & - & 59.31\\
         \hspace{0.75cm}respiration rate & numeric & - & 59.55\\
         \hspace{0.75cm}pulse frequency & numeric & - & 91.91\\
         \hspace{0.75cm}oxygen saturation & numeric & - & 57.18 \\
         \midrule
         \textbf{\Circled{4} contextual information}\\
         \hspace{0.75cm}isolation & discrete & 11 & 1.81\\
         \hspace{0.75cm}transport & discrete & 6 & 59.74 \\
         \hspace{0.75cm}disposition & discrete & 13 & 90.56 \\
         \bottomrule
    \end{tabular}
\end{table}

%% file: sections/5_discussion.tex
\section{Discussion and Limitations}

Our experimental evaluation using both, synthetic and real-world data, provided several insights into the problem domain addressed in this work. First of all, we were able to demonstrate that a correlation-based learning approach for the extraction of disease patterns is indeed capable of identifying meaningful indicators that are closely related to a particular disease under surveillance. 
In particular, the learned definitions showed a similar fit to the real distributions as handcrafted expert definitions (Figure~\ref{fig:comparison}).
Also, the experiments with synthetic syndrome definitions showed a good reconstruction rate, and the discovered real-world syndrome definitions contained plausible features.

Nevertheless,  
the frequent inclusion of unrelated indicators revealed some challenges and limitations of such an approach. Most of them relate to the fact that the training procedure has only limited access to the target information associated with each patient. In contrast to fully labeled data, where information about each patient's medical conditions are available, the learning method is restricted to broad information about a large group of individuals. In addition, the use of temporally aggregated data, depending on its granularity, introduces ambiguity into the learning process. As a result of these constraints, several solutions that satisfy the evaluation criterion to be optimized by the learner exist, even though many of them are undesirable from the perspective of domain experts. This is evident from the fact that the tested algorithm, regardless of the disease and the features used for training, was always able to find strongly correlated patterns, despite the use of unrelated indicators. As another source of problems, we identified the noise, potential inconsistencies and missing pieces of information that may be encountered when dealing with unprocessed and unfiltered real-world data. The consequences become most obvious when taking a look at the results with respect to Norovirus, where the algorithm failed to detect meaningful syndrome descriptions due to the overlap to other, more frequent, diseases with a similar seasonality and more pronounced patterns in the reported numbers.

So far, we were only interested in the identification of patterns the match the target variables as accurate as possible. However, the goal of machine learning approaches usually is to obtain predictions for unseen data. To be able to generalize well beyond the provided training data, this requires models to be resistant against noise and demands for techniques that effectively prevent overfitting. The incorporation of such techniques into our learning approach may improve its ability to find useful patterns despite the noise and ambiguities that are present in the data. For example, successful rule learning algorithms often come with pruning techniques that aim at removing problematic clauses from rules after they have been learned. This requires to split up the training data into multiple partitions in order to be able to obtain unbiased estimates of a rule's quality, independent of the data used for its induction. By splitting up the time series data, the quality of indicators that are taken into account for the construction of syndromes could more reliably be assessed in terms of multiple, independent estimates determined on different portions of the data. Despite such technical solutions, we believe that the active participation of domain experts is indispensable for the success of machine-guided syndromic surveillance. An interactive learning approach, where the syndromes that are discovered by an algorithm are suggested to epidemiologists and feedback is fed back into the system, may prevent the inclusion of undesired patterns and would most likely help to increase the acceptance of machine learning methods among healthcare professionals.

Furthermore, we consider the use of the Pearson correlation coefficient as a limitation of our approach. When modeling the outbreak of a disease, it is especially important to properly reflect the points in time that correspond to high numbers of infections. Other correlation measures, like weighted variants of the Pearson correlation coefficent, may provide advantages in this regard. We expect this aspect to be particularly relevant when modeling rather infrequent diseases with generally low incidences. 
Another problem are possible discrepancies between the data obtained from the emergency departments and the data that incorporates information about the number of infections, e.g., resulting from reporting delays. To circumvent potential issues that may result from such inconsistencies, approaches that have specifically been designed for measuring the similarity between temporal sequences, like dynamic time warping~\citep{muller2007DWT}, could be used in the future. They allow for certain static, and even dynamic, displacements of the sequences to compare.

%% file: sections/6_conclusion.tex
\section{Conclusions}

In this work we have presented a novel approach for the automatic induction of syndrome definitions from health-related data sources. As it aims at finding patterns that correlate with the reported numbers of infections, as provided by publicly available data sources, there is no need for labeled training data. This reduces the burdens imposed on domain experts, who otherwise must manually create labeled data in a laborious and time consuming process. Although the proposed algorithm is able to identify meaningful indicators, due to artifacts in the data and technical limitations, we have found that autonomously created syndromes are likely to include indicators that are unrelated to the disease under surveillance. As a result, the knowledge of experts is still indispensable for the evaluation and supervision of such a machine learning method. Nevertheless, our investigation shows the potential of data-driven approaches to syndromic surveillance, due to their ability to process large amounts of data that cannot fully be understood and analyzed by humans.

In the future, we plan to investigate technical improvements to our algorithm that may help to prevent overfitting and allow for a more extensive, yet computationally efficient, exploration of promising combinations of indicators. In addition, valuable insights can possibly be obtained by applying our approach to different types of health-related data sources, as well as by the investigation of different correlation measures that can potentially be used to guide the search for meaningful syndromes.

\remove{

\subsection{Future Work}

\begin{itemize}
    \item Not only emergency departments. Can be applied on any kind of health-related data source.
    \item Beam search can be used to avoid interesting combinations of indicators being pruned.
\end{itemize}
}

%% file: sections/6_acknowledgments.tex
\section*{Acknowledgments}
We thank our project partners, namely the \emph{Health Protection Authority of Frankfurt}, the \emph{Hesse State Health Office and Centre for Health Protection}, the \emph{Hesse Ministry of Social Affairs and Integration}, the \emph{Robert Koch-Institut}, the \emph{Epias GmbH} and the \emph{Sana Klinikum Offenbach GmbH}, who provided insight and expertise that greatly assisted the research. Especially, we thank Theresa Kocher, Birte Wagner and Sonia Boender from the \emph{Robert-Koch-Institut} for providing the data used for evaluation and for their valuable feedback that substantially improved the manuscript.